\documentclass{article}
\usepackage{arxiv}

\usepackage[utf8]{inputenc} 
\usepackage[T1]{fontenc}    
\usepackage{hyperref}       
\usepackage{url}            
\usepackage{booktabs}       
\usepackage{amsfonts}       
\usepackage{nicefrac}       
\usepackage{microtype}      
\usepackage{lipsum}
\usepackage{graphicx}
\usepackage{amssymb}
\usepackage{amsmath}
\usepackage{csquotes}
\usepackage{mwe}
\usepackage{subfig}
\usepackage{listings}
\usepackage{multirow}
\usepackage{url}
\usepackage{hyperref}
\usepackage{color}
\usepackage{amsthm}%
\usepackage{mathrsfs}%
\usepackage[figuresright]{rotating}%
\usepackage[title]{appendix}%
\usepackage{xcolor}%
\usepackage{textcomp}%
\usepackage{manyfoot}%
\usepackage{algorithm}%
\usepackage{algorithmicx}%
\usepackage{algpseudocode}%
\usepackage{program}%

\newtheorem{assumption}{Assumption}%
\newtheorem{definition}{Definition}%
\usepackage{times}
\usepackage{latexsym}
\algnewcommand{\IIf}[1]{\State\algorithmicif\ #1\ \algorithmicthen}
\algnewcommand{\EndIIf}{\unskip\ \algorithmicend\ \algorithmicif}

\newcommand{\ruparr}{{\color{red}$\uparrow$}}
\newcommand{\gdoarr}{{\color{green}$\downarrow$}}

\usepackage{pifont}
\newcommand{\cmark}{\color{green}\ding{51}}%
\newcommand{\xmark}{\color{red}\ding{55}}%

\definecolor{dkgreen}{rgb}{0,0.6,0}
\definecolor{gray}{rgb}{0.5,0.5,0.5}
\definecolor{mauve}{rgb}{0.58,0,0.82}

\title{Truthful Meta-Explanations for Local Interpretability of Machine Learning Models}

\author{
  Ioannis Mollas\\
  Aristotle University of \\Thessaloniki, 54636, Greece\\
    \texttt{iamollas@csd.auth.gr}\\
     \And
       Nick Bassiliades\\
  Aristotle University of \\Thessaloniki, 54636, Greece\\
             \texttt{nbassili@csd.auth.gr}\\
             \And
               Grigorios Tsoumakas\\
  Aristotle University of \\Thessaloniki, 54636, Greece\\
             \texttt{greg@csd.auth.gr}\\
}

\begin{document}
\maketitle
\begin{abstract}
Automated Machine Learning-based systems' integration into a wide range of tasks has expanded as a result of their performance and speed. Although there are numerous advantages to employing ML-based systems, if they are not interpretable, they should not be used in critical, high-risk applications where human lives are at risk. To address this issue, researchers and businesses have been focusing on finding ways to improve the interpretability of complex ML systems, and several such methods have been developed. Indeed, there are so many developed techniques that it is difficult for practitioners to choose the best among them for their applications, even when using evaluation metrics. As a result, the demand for a selection tool, a meta-explanation technique based on a high-quality evaluation metric, is apparent. In this paper, we present a local meta-explanation technique which builds on top of the \textit{truthfulness} metric, which is a faithfulness-based metric. We demonstrate the effectiveness of both the technique and the metric by concretely defining all the concepts and through experimentation.
\end{abstract}

\section{Introduction}
\label{intro}
Machine Learning (ML), a field of Artificial Intelligence, paves the way for new emerging technologies in a wide variety of sectors, leading the technological advancements. ML is providing solutions to manufacturing; in applications of predictive maintenance~\cite{pm1,pm2}, to banking; with credit scoring~\cite{cs1,cs2} and risk management~\cite{rm1}, as well as to insurance; for fraud detection~\cite{fd1} and damage estimation~\cite{de1}, and, finally, to healthcare; with efficiency of care delivery~\cite{hc1} and diagnosis or prognosis of a plethora of diseases~\cite{hc2,hc3,hc4}.

Even though ML elevates those sectors, societal and ethical issues may arise in high-risk scenarios. For example, credit score predicting algorithms are discriminating between minority and majority populations, leading minorities to poverty and homelessness~\cite{acs1}. As a result of worries about performance, biases and poor trust, an insurance company pulled the plug on an AI tool that was designed to detect fraud in claims through videos~\cite{afd1}. Reports of inappropriate patient treatments~\cite{ahc1}, as well as the use of biased risk prediction models~\cite{ahc2}, have raised concerns in both society and the research community. As a result, legal frameworks and regulations given by many sources, such as the \textit{General Data Protection Regulation} (GDPR)~\cite{gdpr} of the EU and the \textit{Equal Credit Opportunity Act} of the US~\footnote{ECOA 15 U.S. Code \S1691 et seq.}, aim to establish requirements that every ML-powered system should satisfy.

One of the requirements is explainability, which led to the establishment of the Explainable AI (XAI) area~\cite{xais1,xais2}. Interpretable ML (IML), a subfield focused on the interpretations of machine learning models, has attracted the attention of the research community~\cite{imsl1,imls2}. Not every machine learning model, especially deep learning models, can provide interpretations on its own. Since a lot of models are unable to provide interpretations intrinsically, IML has introduced techniques to solve that issue. Such interpretations can be in the form of {\em Feature Importance}, {\em Rules}, and {\em Counterfactual} explanations, among others. Particularly, {\em Feature Importance} (FI) interpretation techniques estimate the influence of each feature on the prediction. Each type of interpretation has its own set of evaluation metrics, and for FI, they include \textit{robustness}~\cite{robustness}, \textit{faithfulness}~\cite{faithfulness}, \textit{infidelity}~\cite{infidelity}, and \textit{truthfulness}~\cite{altruist}.

Researchers and practitioners are experiencing difficulties selecting which of the numerous distinct interpretability techniques introduced is best suited to their application. As a result, an ensemble of interpretability techniques or an automatic selection tool can be extremely valuable. A few ways to address the ensembling of interpretability techniques is through aggregation, for example, via averaging techniques~\cite{DBLP:conf/ijcai/BhattWM20}. However, research on this topic is limited. These methods rely heavily on interpretability metrics. Numerous measures have been proposed to assess the quality of an interpretation, including \textit{fidelity}~\cite{guidotti2018survey}, \textit{coverage}~\cite{guidotti2019factual}, and \textit{stability}~\cite{guidotti2019stability}. Nonetheless, most of them are not useful to the end user.

Explaining the explanations is another interesting direction. Argumentation can be a first step towards this direction. In particular, argumentation is the study of how conclusions can be reached by a logical chain of reasoning, that is, claims based, soundly or not, on premises~\cite{argBook}. IML and argumentation both aim to persuade someone to accept the legitimacy of a decision. In the philosophy of science, it is debatable whether the explanations are arguments or not. An intriguing point of view distinguishes between arguments and explanations, stating that arguments are used to justify something in dispute, but explanations are used to provide a meaning of something incomprehensible~\cite{argExp2}.

In this work, based on our previous preliminary work~\cite{altruist}, we aim to combine 3 concepts to create a meta-explainable ensemble of interpretation techniques based on a complete and user-oriented interpretability metric, complemented with an argumentation framework.

Section~\ref{sec:back} of the paper covers the necessary theoretic concepts, while Section~\ref{sec:relwork} presents related studies. Section\ref{sec:tma} introduces our technique, and Section~\ref{sec:exp} evaluates it through a series of experiments. Finally, in Section~\ref{sec:concl}, we discuss the findings, and provide our concluding opinions and future plans.

\section{Background}
\label{sec:back}
In this section, we introduce the basic notions that underlie our approach. We will discuss machine learning and interpretable machine learning concepts, as well as a few argumentation frameworks.

\subsection{Machine learning}
\label{sec:ml}

Machine Learning (ML) is a cutting-edge technology that forms the core of new and innovative products. We can use ML to solve both supervised and unsupervised problems. In this paper, we emphasise on supervised problems such as binary classification and regression. Thus, given a dataset $D$, containing instances $x_i \in X \subseteq \mathbb{R}^l$, where $l$ is the size of the feature space $F=[f_1, f_2, \dots, f_l]$ and their predictions $y_i \in Y \subseteq \mathbb{R}$, we can train a model $P$ to predict $y$ given an instance $x$, $P(x)=y$. 

Based on the data type $x_i$ can have different shapes. In tabular data, $x_i$ has $l$ values according to the $l$ different $f_i$ features. Dealing with textual data, such as sentences, we can have multiple representations. The simplest representation is to use Bag-of-Words or TF-IDF vectors, which are one-dimensional and express each sentence $x_i$ as a vector with $l$ different values, where $l$ is equal to the size of the vocabulary. We can also have more complex representations, such as word embeddings, which can be two-dimensional. These representations, given a fixed sentence length $s$, express each word of the sentence as a vector of size $e$. If $x_i$ represents multivariate time-series data, then it contains $l\times m$ values, thus $l$ values according to the $l$ different $f_i$ features across $m$ time-steps. Finally, when dealing with images, we must handle with three-dimensional inputs. The first two dimensions represent the image's resolution, while the third expresses its colour channel. Therefore, we can deal with either 1D (tabular or textual data), 2D (textual or time-series data) or 3D (image data) inputs.

We can choose from a variety of ML models according to the task, data type, and size of the dataset, ranging from traditional algorithms (logistic regression and support vector machines) to ensemble algorithms (random forests and XGBoost) and deep neural networks such as CNNs, LSTMs, and Transformers. In this work, we will focus on neural networks and will employ three different types. The first type of neural network will be linear. This network has only an input layer and an output layer. The data, regardless of shape, are handled by the network via a flattening layer. The second type concerns network architectures that are designed specifically for the task, while the third type is a more complex version of them. These networks will contain feed-forward, convolutional, recurrent, bidirectional, and attention layers to showcase our approach to a wide range of networks.

\subsection{Interpretable machine learning}
\label{section:iml}

With the increasing adaptation of ML, there is a need for more transparent and understandable decision systems in a lot of sectors. IML, a subfield of XAI, aims to make ML models more accessible and transparent. There are intrinsically interpretable ML models, like linear models or decision trees, while others, like ensembles or neural networks, most of the time are more complex and uninterpretable. As a result, we require techniques to explain them.

IML approaches might be global, revealing an ML system's whole structure and working mechanism, or local, explaining a specific decision. We can also distinguish between techniques that are applicable to any ML model, known as model-agnostic techniques, and techniques that are limited to specific ML algorithms or architectures, known as model-specific techniques. For example, RuleFit is a global, model-specific technique~\cite{rulefit}, while LIME is a local, model-agnostic technique~\cite{lime}.

Another aspect could be the applicability of an interpretability technique to different data types. There are algorithms that are applicable to specific data types, or there are data-type independent algorithms. For example, LionForests~\cite{mollas2022conclusive} is a data type specific algorithm applicable only to tabular data, while Anchors~\cite{anchors} is a data-type independent algorithm. Furthermore, we can distinguish the difference of the interpretability techniques based on how they provide explanations. There are numerous ways to present an explanation. Several techniques generate rule-based explanations, while others use weights to indicate the importance of input features.

The latter has been expressed using various terms, such as \textit{attribution importance}, \textit{saliency maps}, and \textit{feature importance}, among others. In this work, we will use the last notation, \textit{feature importance} (FI). Depending on the interpretability technique, FI explanations can be global or local. Therefore, given a model $P$, an instance $x_i$ (as presented in Section~\ref{sec:ml}), and an FI interpretability technique $Z$, the interpretation will be $Z(P,x_i)=[z_1, z_2, \dots, z_l]$, where $z_j$ corresponds to a weight -- a.k.a. attribution or importance score -- for the $j^{th}$ value of instance $x_i$.

A variety of algorithms have been proposed in this field. LIME~\cite{lime}, SHAP~\cite{shap}, and Permutation Importance~\cite{rf} are among the most well-known model-agnostic feature importance interpretability algorithms. A plethora of model-specific algorithms, on the other hand, have also been proposed. In neural networks, algorithms exploiting back propagation operation, like Layer-wise Relevance Propagation (LRP)~\cite{lrp} and Integrated Gradients (IG)~\cite{ig}, are retrieving the influence of the input to the output. 

\subsection{Argumentation}
Argumentation theory is a fundamental concept in AI with numerous applications, one of which is in the criminal justice field~\cite{argLaw}. Argumentation procedures show step-by-step how they reached a decision. Therefore, argumentation is considered highly interpretable~\cite{argExp}. However, that's not always the case. Every argumentation procedure is based upon an argumentation framework. Regarding the argumentation framework employed, a few argumentation procedures are interpretable, but not explainable. Classic argumentation based on logic, as proposed by Hunter et al.~\cite{argHunter}, is a simple, yet explainable argumentation framework with many capabilities.

Argumentation based on Classical Logic ($CL$) concerns a framework defined exclusively with logic rules and terms. A sequence of inference to a claim is an argument in this framework. Specifically, an argument is a pair $\langle \Phi, \alpha \rangle$ such that $\Phi$ is consistent ($\Phi \nvdash \perp$), $\Phi \vdash \alpha$, and $\Phi$ is a minimal subset of $\Delta$ (a knowledge base), which means that there is no $\Phi'\subset\Phi$ such that $\Phi' \vdash \alpha$. $\vdash$ represents the classical consequence relation. In this framework counterarguments, the defeaters, are defined as well. $\langle \Psi, \beta \rangle$ is a counterargument for $\langle \Phi, \alpha \rangle$ when the claim $\beta$ contradicts the support $\Phi$. Furthermore, two more specific notions of a counterargument are defined as \textit{undercut} and \textit{rebuttal}. Some arguments specifically contradict other arguments' support, which leads to the undercut notion. An undercut for an argument $\langle \Phi, \alpha \rangle$ is an argument $\langle \Psi, \neg(\phi_1 \land \dots \land \phi_n)\rangle$ where $\{\phi_1, \dots, \phi_n\} \subseteq \Phi$. If there are two arguments in objection, we have the most direct form of dispute. This case is represented by the concept of a rebuttal. An argument $\langle \Psi, \beta \rangle$ is a rebuttal for an argument $\langle \Phi, \alpha \rangle$ if $\beta\leftrightarrow\neg\alpha$.

Argumentation begins when an initial argument is put forward, and some claim is made. This leads to an argumentation tree \textit{Tr} with root node the initial argument. Objections can be posed in the form of a counterargument. In \textit{Tr}, these are represented as children of the initial argument. The latter is addressed in turn, ultimately giving rise to a counterargument. Finally, a judge function decides if a \textit{Tr} is rather Warranted or Unwarranted, based on marks assigned to each node as either undefeated \textit{U} or defeated \textit{D}. A \textit{Tr} is judged as Warranted, \textit{Judge(\textit{Tr}) = Warranted}, if \textit{Mark($A_r$) = U} where $A_r$ is the root node of \textit{Tr} is undefeated. For all nodes $A_i \in$ \textit{Tr}, if there is a child $A_j$ of $A_i$ such that \textit{Mark($A_j$) = U}, then \textit{Mark($A_i$) = D}, otherwise \textit{Mark($A_i$) = U}.

\section{Related Work}
\label{sec:relwork}

In this section, we will present feature importance evaluation metrics found in the literature, as well as a few meta-explanation techniques we identified.

\subsection{Evaluation}
\label{sec:eval_metrics}

One key evaluation metric in the IML research area is \textit{fidelity}. It was first used to evaluate the performance of surrogate models and their ability to mimic the black box models they were explaining. We can define \textit{fidelity} as the accuracy of a surrogate model on a test set in relation to the complex model's decisions. This metric, however, had a number of shortcomings because it was not user-centric and could not be used in non-surrogate interpretation techniques. Influenced by \textit{fidelity}, \textit{faithfulness} and faithfulness-based metrics were therefore introduced~\cite{faith}.

While the origins of the initial Faithfulness-based measure are unclear, one of the first research to propose it aimed at evaluating sentence-level interpretations in text classification tasks~\cite{faithfulness}. In this study, for a given instance, the sentence with the highest important score was removed and the change in the prediction was recorded. The higher the change in the prediction, the better the interpretation. A different definition for \textit{faithfulness} was also provided by a study~\cite{robustness}, measuring the correlation between importance and prediction by continuously removing the most important elements from the input and observing the output.
 
Several variations on \textit{faithfulness} were also introduced. \textit{Decision Flip (most informative token)} removes the most informative token and awards the interpretation if and only if the prediction is changing~\cite{flip1}, whereas Decision Flip (fraction of token) identifies the number of important tokens that must be removed to flip the model decision~\cite{flip2}.

Two other metrics, \textit{comprehensiveness} and \textit{sufficiency}, were introduced as \textit{faithfulness} alternatives~\cite{eraser}. The former evaluates the interpretation by deleting a set of elements from the input and observing the change in the prediction, whereas the latter does so by preserving only the important ones and removing the rest.

\textit{Monotonicity}, also known as \textit{PP Correlation}, is another similar metric~\cite{mono}. It adds components in descending order of priority, beginning with an empty input. The prediction should increase proportionally to the importance of the new elements. The correlation between the prediction and importance scores is then used to calculate \textit{monotonicity}.

\textit{Truthfulness} was introduced as a faithfulness-based metric, which focuses only on the polarity of the feature importance weights~\cite{altruist}. It analyses every element of the input and making different alterations it monitors the model's behaviour. One additional metric, influenced by \textit{faithfulness} and \textit{truthfulness}, proposed to both consider importance correlation and polarity consistency, is \textit{Faithfulness Violation Test} \cite{LiuLGKL022}. This metric captures both the correlation between the importance scores and the change in the probability, while it also examines if the sign of the explanation weights correctly indicates the polarity of input impact, similarly to \textit{truthfulness}.

A few studies introduced datasets with ground truth \textit{rationales}, which are golden explanations. Rationales can be used to evaluate explainability techniques using traditional ML metrics like F1 score and area under the precision-recall curve (AUPRC). One work proposed ERASER, a benchmark for NLP models, which includes datasets containing both document labels and snippets of text recognized as explanations by annotators~\cite{eraser}. However, in real applications, most of the datasets do not contain ground truth information regarding the explanations, and as such these evaluation approaches cannot be applied. Furthermore, we can only assume that humans are capable of annotating unbiased \textit{rationales}~\cite{tan22}. Nevertheless, the usefulness of such benchmarks is to enable comparison of newly proposed interpretability techniques.

\subsection{Meta-explanations / Aggregation}

Different aggregation procedures are initially introduced in a very interesting research~\cite{DBLP:conf/ijcai/BhattWM20}. Attempting to combine multiple explanation techniques, metrics such as \textit{sensitivity}, \textit{faithfulness}, and \textit{complexity}, are used over different combination strategies. Among these combination strategies, Mean and Median, are presented. The results suggest that aggregating results to a smaller typical error, compared to the error an explanation by one technique can have. Moreover, another combination strategy is presented. For a given instance, a set of near neighbours is identified. Extracting explanations for the predictions of these neighbours, the final explanation is the aggregation of the explanations, weighted by the distance of the neighbours to the original instance. The latter was designed to lower the \textit{sensitivity} and \textit{complexity}.

A recent research introduced a method that enables the combination of explanations provided by multiple techniques, using specific evaluation metrics, to do so~\cite{bobek2021towards}. In their experiments, they use LIME, SHAP and Anchors to ensemble explanations. They select three metrics; \textit{stability}, \textit{consistency}, and \textit{area under the loss curve}, to ensemble the weights produced by the techniques into one. One issue with this approach is that the \textit{consistency} metric requires to create and use different ML models to produce explanations. One of the framework's shortcomings is that it only enables model weighting using comparative evaluation metrics across several models/explainers. It does not guarantee that the final explanations are correct or acceptable for the end user. This method was evaluated only in an image classification use case.

Finally, another work on ensembling explanations introduces EBEC, a method for correcting global explanations of non-differentiable ML models with a non-differentiable importance score~\cite{hamamoto2021model}. The central idea of EBEC is to train multiple ML models on a dataset to identify different local minima, then produce global interpretations using an explainability technique (in this work SHAP), and finally combine them by solving an optimization problem that guarantees certain qualitative properties. They conclude that EBEC works effectively in three different tabular datasets based on their evaluation.

\section{Truthful meta-explanations supported by arguments}
\label{sec:tma}
In this work, we are presenting a three-dimensional contribution to the IML community. Focusing exclusively in FI interpretation techniques, we first formulate the \textit{truthfulness} definition. Then, we present methods for aggregating multiple interpretations in an ensemble fashion. Finally, we also present a meta-explain process, which uses the \textit{truthfulness} metric. 

To begin, we will state a few assumptions that must hold for our technique to be theoretically sound. Assumption~\ref{assumption:1} ensures that the ML model we are trying to apply our technique is able to provide continuous predictions. This is a necessary property for the metric we are going to formulate in the following section (Section~\ref{ssec:truth}).

\begin{assumption}
\label{assumption:1}
The machine learning model $P(x) = y$ can provide continuous predictions $y \in \mathbb{R}$. A classification model, for example, should be able to provide predictions in the form of probabilities. A regression model, on the other hand, always produces continuous outputs.
\end{assumption}

The second assumption (Assumption~\ref{assumption:2}) concerns the interpretability techniques utilized in the approach. The amount and type of the technique to be used in the ensemble to produce one final interpretation is not limited, with the only exception being to provide weights that are either capture presents a weight for a feature which represents a (local) monotonic relation to the prediction of a specific label or being perceived by end users as such. This is critical since only a few interpretability techniques, such as SHAP, provide the contribution of a feature to the prediction without assuming any local or global monotonicity. Nonetheless, based on this proposed contribution, end-users perceive the relationship between a feature and the output to have monotonic behaviour when altered.

\begin{assumption}
\label{assumption:2}
FI's are producing, or they are perceived as producing, $z_j$ weights with local or global monotonic notion. 
\end{assumption}

\subsection{Truthfulness metric}
\label{ssec:truth}

The first contribution of this paper concerns the \textit{truthfulness} metric. Truthfulness is a user-inspired evaluation metric that simulates a user's behaviour with respect to an interpretation. It addresses issues of other Faithfulness-based metrics by evaluating all feature importance elements and taking into account all signs (Positive, Negative and Neutral). But, before we get into the definition of the metric, we will present an example. The interpretation that follows explains a prediction concerning a user's loan disapproval in a bank. 

\begin{figure}[ht]
  \centering
  \includegraphics[width=0.9\linewidth]{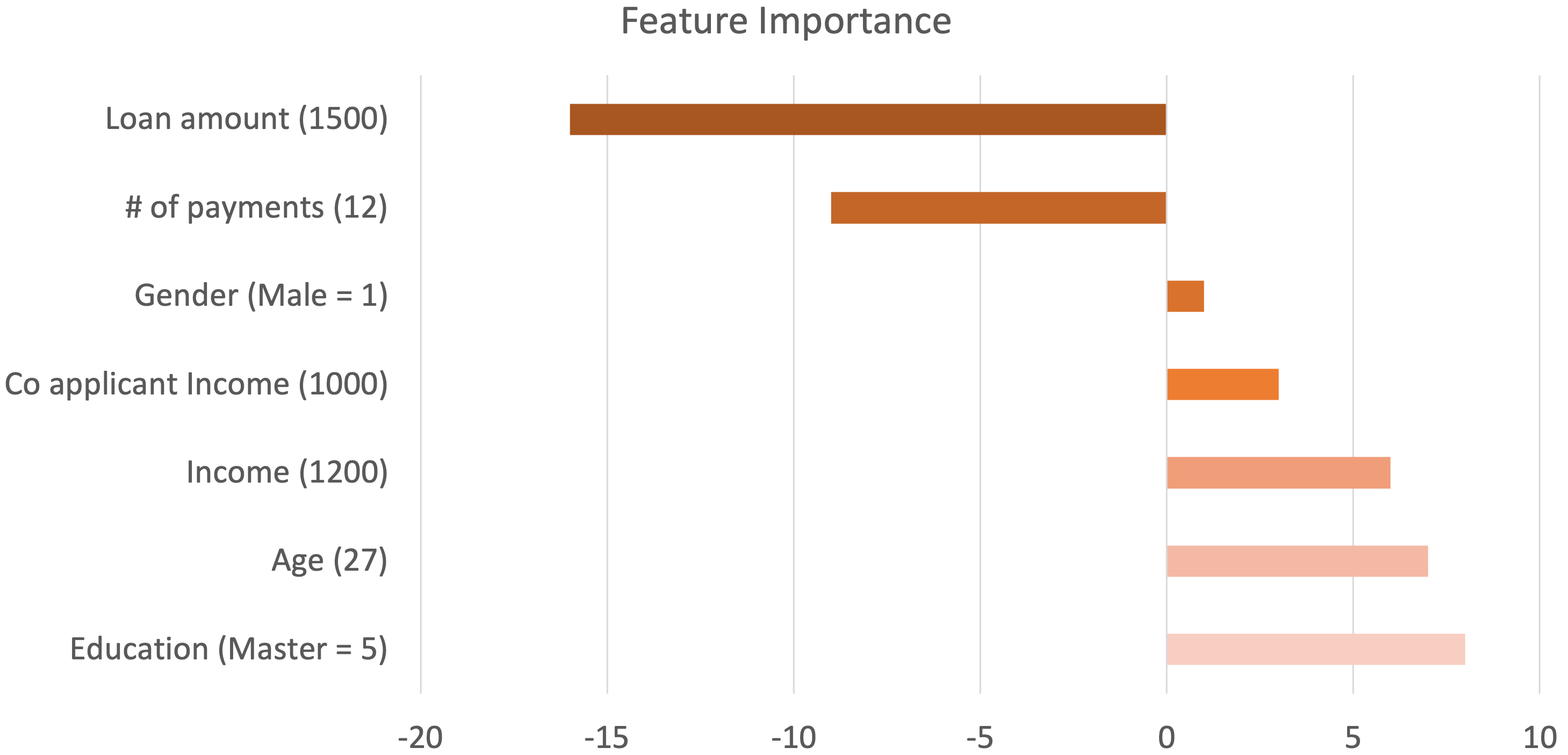}
  \caption{Feature importance weights assigned to the features of the example}
  \label{fig:example_user}
\end{figure}

A user received the interpretation presented in Figure~\ref{fig:example_user}. Her friend's income was \$1,000 (Co applicant Income). She noticed that this had little influence on her decision, resulting in a disapproval with a score of 70 (minimum acceptable 75). As a result, she decided to include her mother, who has a little greater income (\$1.2K), as a co-applicant. To her astonishment, this had no effect on the outcome, as she obtained a score of 70, which she expected to increase slightly. Therefore, the user judged the interpretation as untruthful, while she did not trust the predictive system either.

Influenced by this, we suggest a metric that, given an interpretation, performs a few tests to ensure that the interpretation provided to the end-user is truthful. Hence, for each feature importance score $z_j$ assigned to the feature values $v_i^j$ of $x_i$, we both increase and decrease the feature values, and we observe if the model behaves as expected with respect to the feature importance. 

In this work, we will focus on four types of data: textual, image, tabular, and time-series. The words are the features in textual tasks, and the importance concerns a word and, in some cases, its position. When dealing with images, feature importance is used to describe either a single pixel or a group of pixels known as superpixels. Each feature in tabular data has its own importance score. Lastly, in time-series, feature importance can refer to either a sensor's time-step value or a sensor throughout the entire time-window. In all these cases, feature importance can be either Positive, Negative, or Neutral, as described in Definition~\ref{th:imp}.

\begin{definition}
\label{th:imp}
The importance assigned to a feature can be IMP $\in$ \{$1=$ Positive ($z_i>0$), $-1=$ Negative ($z_i<0$), or $0=$ Neutral ($z_i=0$)\}.
\end{definition}

Let's discuss now how we alter a feature value $v_i^j$. Given a set of samples $X'$, we measure each feature's distribution statistics, namely, min, max, mean ad STD values. Then, as presented in Algorithm~\ref{alg:altvalues}, we calculate a $noise$ or the alternative values. This $noise$ is small, and therefore these alterations are local. This procedure is different for the various data types. In Textual datasets, this procedure replaces the word regarding the examined feature importance score with an empty string. In images, we compute a $noise$ which makes lighter and darker a pixel or a superpixel. For superpixels, we have to also employ an image segmentation algorithm to identify the superpixels of an image. Regarding tabular data, we create a $noise$ which both increases and decreases the feature value, while in time-series, we increase and decrease with the calculated $noise$ either a specific time-step of a sensor, or the whole time-window of a sensor. We set three different $noise$ levels; ``weak'', ``normal'', and ``strong'', in the cases which are applicable (image, tabular and time-series).

\begin{algorithm}
\caption{Process of determining the alternative values for a feature}
\label{alg:altvalues}
\begin{algorithmic}
\Require Instance's feature value $value$, Distribution statistics of feature $feature\_distribution$, $Noise$ level $level$, Data type $type$
\Procedure{DetermineAltValues}{$value$, $feature\_distribution$, $level$, $type$}
\If{$type$ is $Textual$}
    \State $value^+ \gets value$
    \State $value^- \gets 0$
    \State \Return $value^-, value^+$ 
\EndIf
\State $min, max, mean, std \gets extract(feature\_distribution)$
\State $noise \gets abs(mean-gaussian\_noise(mean, std))$
\If{$type$ is $Image$ or $Time Series$}
    \State \Return $-noise, +noise$
\EndIf

\State $value^- \gets value - noise$
\State $value^+ \gets value + noise$

\IIf{$value^- < min$}
    $value^- \gets min$
\EndIIf
\IIf{$value^+ > max$}
    $value^+ \gets max$
\EndIIf
\State \Return $value^-, value^+$
\EndProcedure
\end{algorithmic}
\end{algorithm}

\begin{definition}
\label{th:alt}
The alteration of the value of a feature can be ALT $\in$ \{$1=$ Increasing ($v_{j,i}'>v_{j,i}$), $-1=$ Decreasing ($v_{j,i}'<v_{j,i}$)\}, where $v_{j,i}'$ the altered value.
\end{definition}

\begin{figure}[ht]
  \centering
  \includegraphics[width=0.9\linewidth]{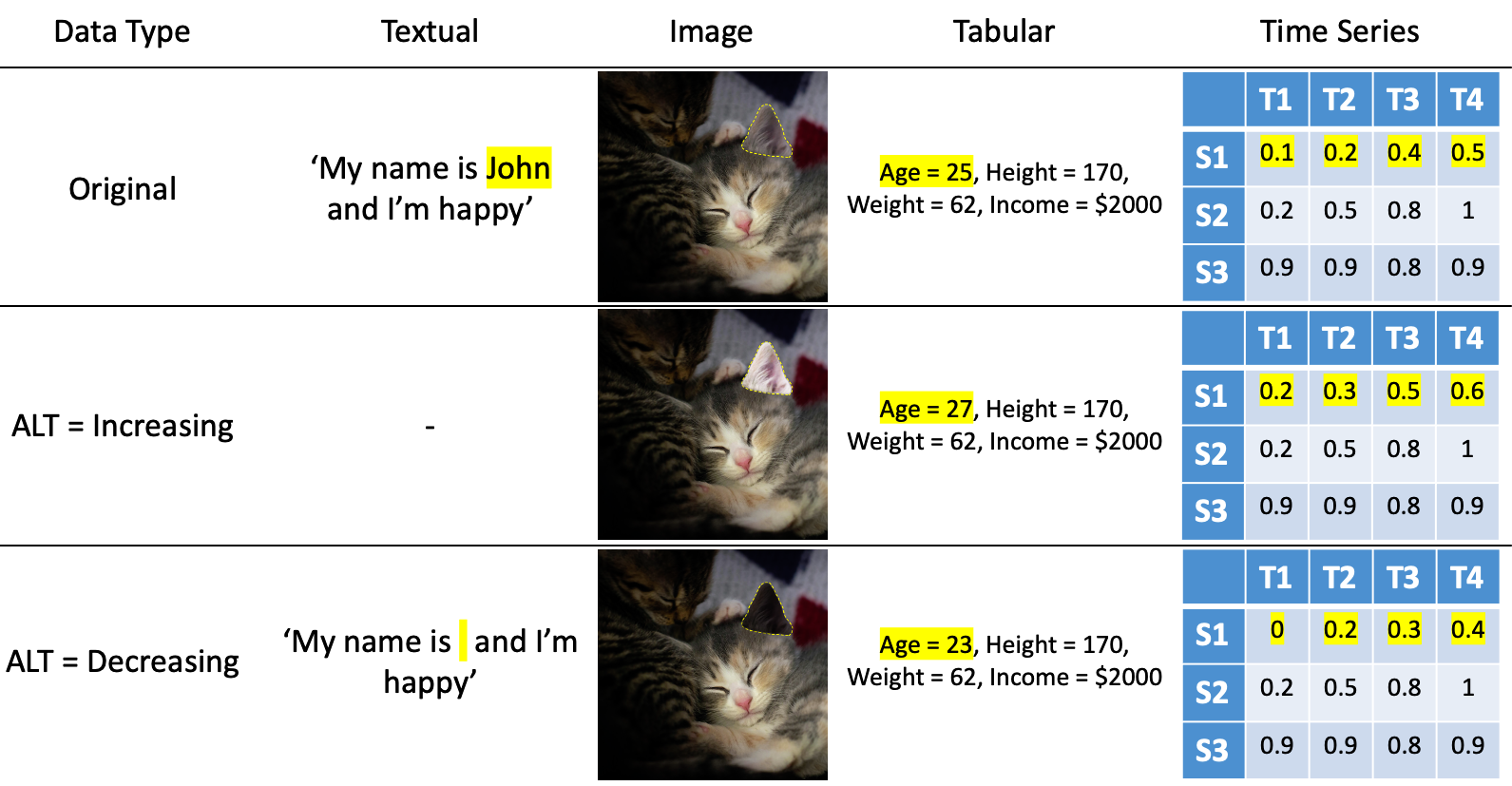}
  \caption{Example of altering a feature of an instance for the four different data types}
  \label{fig:example_alteration}
\end{figure}

In Figure~\ref{fig:example_alteration}, we show an example of each data type. The textual example demonstrates how to remove (decrease) a feature, in this case, the word ``John''. In the image example, we can see that the first alteration involves lightening the kitten's ear by increasing the values of the superpixel, and the second alteration involves darkening the ear by decreasing the values of the pixels. In the tabular example, we make two changes to the ``Age'' feature. We increase ``Age'' from 25 to 27, while also decreasing it to 23. Finally, in the time series, we increase the first sensor's readings by 0.1 at each time step and decrease them by the same value.

We discussed a feature's feature importance score and introduced the concept of alterations in the values of features across different data types. We are now introducing the concept of expected behaviour. Given a feature importance score $z_j$ for $f_j$, we have two alternative values for that feature for a specific instance $x_i$. We can request the ML model to predict the modified instance $x_i' \in \{x_i^{inc}, x_i^{dec}\}$, where $x_i^{inc}$ is the same instance but with a higher value for the examined feature and $x_i^{dec}$ has a lower value. Then, with regard to the feature importance score $z_j$, we evaluate whether the model's predictions $P(x_i^{inc})$ and $P(x_i^{dec})$ behave as expected.


\begin{definition}
\label{th:exp}
The expected behaviour of an \textit{M} component can be EXP $\in$ \{$1=$ Increasing ($P_{M}(x_i) - P_{M}(x_i') < \delta$), $-1=$ Decreasing ($P_{M}(x_i) - P_{M}(x_i') > -\delta $), $0=$ Remaining Stable ($\lvert P_{M}(x_i') - P_{M}(x_i)\rvert < \delta$)\}, where $x_i'$ the instance with the altered value, while tolerance $\delta$ is defined either manually by the user or is set to a default value ($0.0001$).
\end{definition}

As presented in Definition~\ref{th:exp}, the model's prediction can behave in three ways. It can increase, decrease, or remain stable. If the feature importance score $z_j$ is positive, we expect the prediction to increase for the $x_i^{inc}$ modified instance while decreasing for the $x_i^{dec}$. If $z_j$ is negative, we expect the prediction of the two changes, $x_i^{inc}$ and $x_i^{dec}$, to decrease and increase, respectively. In the case of a neutral feature importance score $z_j$, we anticipate that the prediction will remain stable for both alterations. We also use a $\delta$ tolerance value. This will help evaluate an importance score as truthful in cases where the prediction will change from $0.75$ to $0.7502$, where the difference is extremely small. This will help to not punish small mistakes. However, setting $\delta=0$ leads to a stricter evaluation. In the experiments section (Section~\ref{sec:exp}), we examine different delta values. Table~\ref{tab:truth} summarizes all of these.

\begin{table}[ht]
\centering
\caption{Truthfulness matrix [(t)ruthful and (u)ntruthful states]}
\label{tab:truth}
\begin{tabular}{ccccc|cccc}
\cline{3-8}
&    & \multicolumn{3}{c|}{1} & \multicolumn{3}{c}{-1} & \multicolumn{1}{c}{\textbf{ALT}} \\
                                           &    & 1     & 0     & -1     & 1      & 0     & -1     & \multicolumn{1}{c}{\textbf{EXP}} \\ \cline{3-8}
\multicolumn{1}{c}{} & 1  & t     & u     & u      & u      & u     & t      &                                   \\
\multicolumn{1}{c}{{\textbf{IMP}}}                    & 0  & u     & t     & u      & u      & t     & u      &                                   \\
\multicolumn{1}{c}{}                     & -1 & u     & u     & t      & t      & u     & u      &                                   \\ \cline{3-8}
\end{tabular}
\end{table}

As a result, we argue that a feature importance score is truthful if and only if the behaviour of the model's prediction regarding the alterations is as expected. This is also included in Definition~\ref{th:tru}. It is worthwhile to provide an example to demonstrate this. The ML model predicts $P_{M}(x_i)=0.7$ for a random instance $x_i$, and the feature $f_1$, with a value of $v_{1,i} = 1$, has an IMP $z_1 = 0.5$ (positive). We use Gaussian noise to increase and decrease the value of the feature based on its distribution. We change the value to $v_{1,i}^{inc} = 1.21$ and $v_{1,i}^{dec} = 0.85$, for $x_i^{inc}$ and $x_i^{dec}$, respectively. Then, we observe the model's predictions by querying the ML model. In this example, the prediction for the $v_{1,i}^{inc}$ was increased to $0.85$, while the prediction for the $v_{1,i}^{dec}$ remained stable. As a result, we can conclude that the behaviour in the second alteration was not as expected, and hence the feature importance score is untruthful.

\begin{definition}[Truthfulness]
\label{th:tru}
The importance assigned to a feature can be defined as \textit{truthful} when the expected changes to the output of the \textit{M} model $P_{M}(x_i')$ are correctly observed with respect to the alterations that occur in the value of this feature. Thus, for both values of ALT and a given IMP, the IMP$\times$ALT=EXP must be in accordance with the \textit{truthfulness} matrix (Table~\ref{tab:truth}).
\end{definition}

Given an interpretation with feature importance scores for each feature, \textit{truthfulness} as a measure will analyse each score individually and penalize those who were untruthful. Thus, for an instance of $\lvert F\rvert$ features, we examine their importance scores, and for the ones that are truthful we increase the \textit{truthfulness} score by one, while for untruthful scores we do not increase the score. Finally, we divide the \textit{truthfulness} score with the number of the features to normalise the score to $[0,1]$. The aforementioned is mathematically formulated as follows: 

\begin{equation}
    Truthfulness(Z(P,x_i)) = T(Z(P,x_i)) = \frac{1}{\lvert F\rvert}\sum_{j=1}^{\lvert F\rvert}evaluate(z_j,x_i)
\end{equation}

\[
  evaluate(z_j,x_i) =
  \begin{cases}
        \text{1} & \text{if } z_j \text{ truthful with respect to the}\\
                 & \text{alterations of } x_i, x_i' \in \{x_i^{inc}, x_i^{dec} \}\\
        \text{0} & \text{else}
  \end{cases}
\]

Before proceeding with the meta-explanation ensembling technique, we are arguing on why we only use two alternative values to evaluate a feature importance score of a feature for a given instance. We assume that if there is monotonicity between these two values, there will be monotonicity in the intermediate values as well. We make this assumption to minimise the computational cost, considering that many interpretation techniques have high response times. We achieve three things with this choice:
\begin{itemize}
    \item [] \textbf{Faster evaluation:} Given a set of interpretability approaches and their response times, \textit{truthfulness} has a low computing cost when applying two alterations per feature, as opposed to more.
    \item [] \textbf{Integration to a meta-explanation technique:} Being lighter computationally, we can use this metric in a meta-explanation technique to produce even better explanations. In the following section, we introduce a meta-explanation technique that makes use of the \textit{truthfulness} metric.
    \item [] \textbf{Reduce the environmental impact:} Given that \textit{truthfulness} necessitates re-querying the ML model twice for each feature, utilising a set of alternative values rather than two will increase the cost exponentially in larger feature sets. As a result, we anticipate a lower environmental impact by selecting only two alternative values.
\end{itemize}

\subsection{Meta-explanation technique}
\label{sec:ourMeta}
Based on the \textit{truthfulness} metric, we introduce a meta-explanation technique. Differently than the other recent research, we employ \textit{truthfulness} metric to combine multiple explanation techniques, to provide a more accurate explanation. 

Given $X$ interpretation techniques, and an examined instance $x_e$ whose prediction made by a model $M$, $P_M(x_e)$, to ensemble the different interpretations $Z = [Z_0, Z_1, \dots, Z_X]$, we calculate the \textit{truthfulness} of each interpretation. We also measure the average change of the output given the two alterations $x_i'$ of each feature $f_j$, $ac_j = \frac{1}{2}(\lvert P_M(x_i)-P_M(x_i^{inc})\rvert+\lvert P_M(x_i)-P_M(x_i^{dec})\rvert)$.

The first step in performing the ensembling of the multiple interpretations is to determine the candidate importance scores for each feature based on its \textit{truthfulness}. Algorithm~\ref{alg:candidate_fis} illustrates this. For each feature, we verify the \textit{truthfulness} of the importance scores assigned by the various interpretation techniques and save them for use in the next step.

\begin{algorithm}
\caption{Identification of candidate truthful feature importance scores}
\label{alg:candidate_fis}
\begin{algorithmic}
\Require Interpretations $Z$, Examined Instance $x_e$, Feature Set $F$
\Procedure{CandidateTruthfulScores}{$Z$, $x_e$, $F$}
\State $CZ \gets$ new Hash Map 

\For{feature $f_j \in F$}
    \State $temp \gets []$
    \For{interpretation $Z_m \in Z$}
        \State $z_j = Z_m[f_j]$
        \IIf{evaluate($z_j,x_e) = 1$}
            \State $temp \gets temp \cup z_j$
        \EndIIf
    \EndFor
    \State insert $temp$ list in $CZ$ with $f_j$ as key
\EndFor
\State \Return $CZ$
\EndProcedure
\end{algorithmic}
\end{algorithm}

By identifying the truthful importance scores from each interpretation for each feature for an examined instance, we present our ensemble procedure in Algorithm~\ref{alg:ensembling}. First, we sort the features by average change. Then, starting with the feature with the greatest absolute change, we examine the candidate importance scores and choose the one with the highest absolute value. We save the highest absolute value, which we will use in the following steps. We proceed on to the next feature, which has the second largest absolute change. Again, we choose the highest absolute importance score among the candidate scores, which has to be lower than the previous score by absolute value (line 13th). We do this because one technique may have a truthful score but an inappropriate magnitude (e.g., the importance is 1, while it should have been 0.4). We assign a zero value when a feature has no candidate feature importance scores.

\begin{algorithm}
\caption{Truthfulness-based meta-explanation algorithm}
\label{alg:ensembling}
\begin{algorithmic}
\Require Candidate Imp. Scores $CZ$, Average Change $AC$, Feature Set $F$
\Procedure{TruthfulMetaExplanation}{$CZ$, $AC$, $F$}

\State $OF \gets $ sort($F$) based on $AC$, $temp\_score \gets None$
\State $meta\_explanation \gets$ new Hash Map
\For{feature $f_j \in OF$}
    \If{$CZ[f_j]$ is not empty}
        \If{$temp\_score$ is $None$}
            \State $ind \gets$ index of $max\_abs(CZ[f_j])$
            \State $temp\_score \gets max\_abs(CZ[f_j])$
        \Else
            \If{$CZ[f_j]$ has only one value}
                \State $ind \gets 0$
                \State $temp\_score \gets min(abs(CZ[f_j][ind]),temp\_score)$
            \Else
                \State $index \gets$ index of the most appropriate value
                \State $temp\_score \gets min(abs(CZ[f_j][ind]),temp\_score)$
            \EndIf
        \EndIf
        \State insert $CZ[f_j][ind]$ in $meta\_explanation$ with $f_j$ as key
    \Else
        \State insert $0$ in $meta\_explanation$ with $f_j$ as key
    \EndIf
\EndFor

\State \Return $meta\_explanation$
\EndProcedure
\end{algorithmic}
\end{algorithm}

Let's discuss an example using this algorithm. In Table~\ref{tab:example}, we have an instance $x_e = [0, 1, 0.5, 0.3, -0.2]$, and three interpretations $Z = [Z_0, Z_1, Z_2]$ regarding its prediction. We calculate the \textit{truthfulness} of each interpretation. The truthful importance scores of each technique are represented with a green check mark in Table~\ref{tab:example}, while untruthful scores with red cross mark. The average change (AC) of the output for each feature based on two alterations is also provided. Based on the Algorithm~\ref{alg:ensembling}, we select first the score of $f_2$, which has the highest AC score ($ac_2=0.8$). Among the three truthful importance scores, we select the highest $z_2^0=1$, the one from $Z_0$. Then, we proceed to the next feature $f_3$. There is only one truthful importance score $z_3^2=-0.4$ provided from $Z_2$. In the same fashion, we are selecting the most appropriate importance scores for each feature, till we have a complete interpretation.

\begin{table}[ht]
\centering
\caption{Example of evaluation of three techniques and the meta-explanation}
\label{tab:example}
\begin{tabular}{cccccc}
\hline
       & $f_1$      & $f_2$      & $f_3$      & $f_4$       & $f_5$      \\ \hline
$Z_0$  & 0\cmark    & 1\cmark    & 0.5\xmark  & 0.3\xmark   & -0.2\cmark \\
$Z_1$  & 0.1\cmark  & 0.23\cmark & 0.7\xmark  & 0.3\xmark   & 0.3\xmark  \\
$Z_2$  & -0.1\xmark & 0.2\cmark  & -0.4\cmark & 0.2\xmark   & -0.1\cmark \\ \hline
$AC$   & 0.05       & 0.8        & 0.5        & 0.01        & 0.3        \\ \hline
$meta$ & 0.1        & 1          & -0.4       & 0           & -0.2       \\ \hline
\end{tabular}
\end{table}

With this ensembling procedure, we achieve two things. The first is that we gather more truthful importance scores in the final interpretation. Moreover, we re-rank the importance scores of the features using the \textit{truthfulness} evaluation and our ensembling algorithm. Later in the experiments, we will discuss the effectiveness of our approach.

\subsection{Argumentation}

The argumentation framework we designed to provide justifications for the \textit{truthfulness} evaluation was a very useful component of Altruist, our earlier preliminary work. While we do not re-formulate the entire argumentation system, in this section, we do re-formulate the atoms which form arguments, utilised in our system to make them more descriptive. More information about the theoretical formulation of the framework can be found in our earlier work~\cite{altruist}. The original available atoms were the following:

\begin{itemize}
    \item []$a$: The explanation is untrusted
    \item []$b$: The explanation is trusted
    \item []$c_j$: The importance $z_j$ is untruthful
    \item []$d_j$: The importance $z_j$ is truthful
    \item []$e_{j,ALT}$: The model's behaviour by altering $f_j$'s value is not according to its importance
    \item []$f_{j,ALT}$: The evaluation of the alteration of $f_j$'s value was performed and the model's behaviour was as expected, according to its importance.
\end{itemize}

We are re-phrasing the last two atoms, $e_{j,ALT}$ and $f_{j,ALT}$, as seen below:

\begin{itemize}
    \item []$e_{j,ALT}$: The model's behaviour by altering $f_j$'s value from $X$ to $Y$ ($ALT$) is not according to its importance $Z$
    \item []$f_{j,ALT}$: The evaluation of the alteration of $f_j$'s value $X$ to $Y$ ($ALT$) was performed and the model's behaviour was as expected $EXP$, according to its importance $z_j$.
\end{itemize}

This way, we do not modify the theoretic argumentation framework supporting our system, but we are making the last two atoms used in the arguments more descriptive. While we are presenting a complete example in the qualitative experiments, we are showing an example below:

\begin{itemize}
    \item []$e_{2,INC}$: The model's behaviour by altering $f_2$'s value from $25$ to $26$ (increased) is not according to its importance $Z$
    \item []$f_{2,INC}$: The evaluation of the alteration of $f_2$'s value $25$ to $26$ (increased) was performed and the model's behaviour was as expected (increased), according to its importance $z_2$.
\end{itemize}

An example showcasing the enhanced argumentation framework is being presented in the qualitative experiments (Section~\ref{sec:argExa}).

\section{Experiments}
\label{sec:exp}
In this section, we will test the \textit{truthfulness} metric on several types of datasets, as well as our meta-explanation technique, in a series of quantitative experiments. We will also conduct a qualitative evaluation of the explanations produced by the meta-explanation technique.

\subsection{Setup}
We will begin by describing the datasets we used, the preprocessing procedures we utilised, the predictive models we employed, and the interpretation techniques we included in our experiments.

\subsubsection{Datasets}
We included the following datasets in our experiments in order to cover a variety of the critical sectors that use ML in their workflows, as presented in Section~\ref{intro}. We incorporated the Turbofan Engine Degradation Simulation (TEDS) dataset~\cite{saxena2008turbofan,saxena2008damage} for the manufacturing sector's predictive maintenance scenario, which aims to predict the remaining useful lifetime (RUL) of engines using \emph{time-series data}. The second dataset we are using, Credit Card Approval Prediction (CCA), contains information about bank customers (\emph{tabular data}), as well as information regarding their debt payments (if any)\footnote{\url{https://cutt.ly/xQ1mqyo}}. The goal is to determine whether a client is eligible for a credit card. A dataset for Hurricane Damage Estimation (HDE)\footnote{\url{https://cutt.ly/kQ1n3kE}} of properties using satellite \emph{images}~\cite{sdad}, in a classification manner, is incorporated in our experiments to cover the insurance sector. Finally, data for the identification of Acute Ischemic Strokes (MedN) through brain MRI reports (medical notes - \emph{text})\footnote{\url{https://cutt.ly/sQ1nM9c}} \cite{ais} connects the experiments to the healthcare sector. More information about the datasets is visible in Table~\ref{tab:datasets-info}, while about their preprocessing in Section~\ref{sec:preprocessing} and in the GitHub repository ``TMX: Truthful Meta Explanations''~\footnote{\url{https://github.com/iamollas/TMX-TruthfulMetaExplanations.git}}.

\begin{table}[ht]
\centering
\caption{Information about the datasets incorporated in our experiments. *After preprocessing (R: Regression, BC: Binary Classification)}
\label{tab:datasets-info}
\begin{tabular}{rcccc}
\hline
\textbf{Name} & \textbf{\# of Instances} & \textbf{Task}  & \textbf{Sector} & \textbf{Data Type} \\ \hline
TEDS        & 33.727*  & R  & Manufacturing  & Time-Series        \\
CCA         & 1.232*   & BC & Banking        & Tabular            \\ 
HDE         & 23.000   & BC & Insurance      & Images             \\ 
MedN.       & 3.204    & BC & Health         & Textual            \\ \hline
\end{tabular}%
\end{table}

\subsubsection{Preprocessing}
\label{sec:preprocessing}
While extended preprocessing is accessible in our GitHub repository, here we mention few crucial preprocessing steps. We suggest other researchers and users to apply similar preprocessing towards more interpretable end-to-end systems.

Starting with the time-series dataset, TEDS, we scale our data to $[0.1,1]$ after reducing the available features and retaining only measurements from 14 sensors. We chose to scale our data to this range rather than $[0,1]$ because none of the 14 sensors have measurements that are equal to $0$, but only positive values. This is a critical decision in terms of interpretability. A lot of the time, the $0$ value has a neutral notion in terms of interpretability and, more precisely, feature importance. As a result, local techniques such as LIME can generate a positive weight that, when multiplied by the $0$ value, is neutralised. Furthermore, we create examples of 14 measurements across 50 time steps, using the RUL value on the most recent time step as the goal variable, utilizing a time window of 50 time steps. One final preprocessing step we use is to scale the output, the RUL value, to $[0,1]$, which is easier for a neural network to learn.

In our tabular data, on the other hand, there exist features with both positive, negative, and zero values. In this case, we'd like to scale them to $[-1,1]$ while keeping the centre at zero. To address this issue, we use maximum absolute value scaling. In terms of image preprocessing, we did augmentation by randomly flipping and rotating the samples and scaling them from $[0,255]$ to $[0,1]$. In terms of the textual dataset, we used a symbol-removal process on each document and reduced the maximum number of words from $380$ to $250$, because relatively few documents were longer than $250$ words. Finally, we used the BioBERT~\cite{lee2020biobert} pretrained transformer to obtain word-level embeddings for each document.

\begin{figure}[ht]
  \centering
  \includegraphics[width=0.7\linewidth]{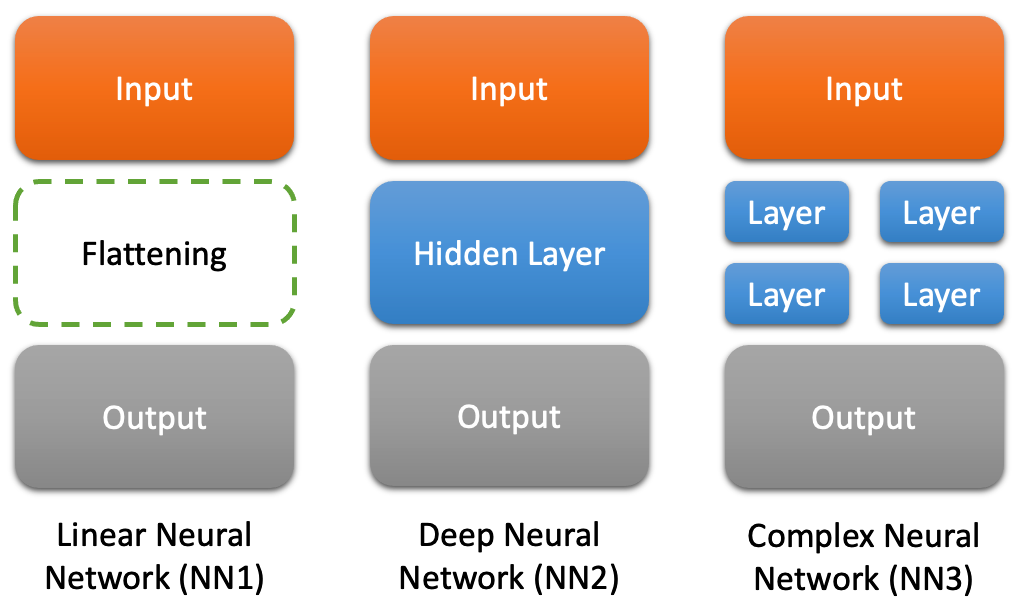}
  \caption{The three different architecture formats used in our experiments}
  \label{fig:arch_formats}
\end{figure}

\subsubsection{Model architectures}

We employed three distinct model architecture formats. The first type of network is a linear network (NN1). We have only input and output layers for all data kinds, but just one additional layer for flattening 2D or 3D data. The second type of architecture is a deep neural network that has been tailored to the various data types (NN2). For example, we used recurrent and feedforward layers for the time series dataset, solely feedforward layers for the tabular data, convolutional and feedforward layers for the image dataset, and bidirectional recurrent, one-dimensional convolutional, and feedforward layers for the textual dataset. The third architecture type is a more complicated variant of the second, with more layers, more neurons, and different activation functions (NN3). We could argue that these networks are ``unnecessary'' complex, but they are required to enable our studies. All these different architecture formats are presented in Figure~\ref{fig:arch_formats}.

\begin{table}[ht]
\centering
\caption{The three different architecture formats used in our experiments}
\label{tab:perf-table}
\begin{tabular}{cccccc}
\hline
Dataset & Metric & Avg \# of Features & NN1 & NN2  & NN3 \\ \hline
TEDS    & RMSE   & 14             & 36.98  & 35.50 & 32.71   \\
CCA     & $F_1$  & 12                 & 58.9\%  & 70.71\% & 71.08\%\\
HDE     & $F_1$  & 29.71              & 56.55\%  & 93.22\% & 85.29\%\\
MedN    & $F_1$  & 48.25              & 91.27\%  & 96.64\% & 95.73\%\\ \hline
\end{tabular}
\end{table}

Table~\ref{tab:perf-table} also shows the performance of each model on the four separate tasks. It is worth noting that the linear neural (NN1) model is consistently outperformed by the other models. However, we include this model in order to evaluate our metric. In all cases, NN2 performs as well as or better than NN3. The majority of neural networks in use are NN2. Finally, NN3 does not outperform NN2 in all circumstances (see datasets HDE and MedN), but such complex models are required to stress out the interpretability techniques based on gradients.

\subsubsection{Interpretation techniques}

In our experiments, we used three different interpretability techniques, one model-agnostic and two model-specific (neural-specific). These are LIME, IG, and LRP, which we discussed in Section~\ref{section:iml}. We use the Mean and the Median for meta-explanation techniques as long as our proposed technique, as presented in Section 4. The Mean meta-explanation technique for each feature computes the mean value across the importance scores provided by the different interpretation techniques, while Median takes the median value. For example, if LIME, IG, and LRP provided 0.2, 0.3, and 0.7 importance scores for the feature ``age'', Mean will assign a 0.4 importance score to ``age'', while Median will assign a 0.3. Finally, we utilise a random approximation of the interpretation as a baseline.

One more interpretation technique would be to exploit the real weights of the linear neural models (NN1). Those weights are ground truth interpretations. However, we will treat our NN1 as a black box model, and we will use these ground truth interpretations to evaluate our metric. Our metric should provide a perfect score for these interpretations, as they are real. 

We will apply these interpretability techniques in all datasets. Regarding TEDS, we will both apply them in time-step and sensor level, where in the former feature importance scores will be assigned to each time-step of each sensor, while in the latter, importance scores are assigned in the time-steps. For the CCA dataset, each feature is assigned a feature importance score, while in HDE, each pixel. Finally, in MedN dataset, the interpretation techniques assigned importance scores to each term (word). 

\subsection{Quantitative experiments}
The quantitative experiments we conducted include the evaluation of the \textit{truthfulness} metric using ground truth information from the NN1, and a comparison of the different interpretability techniques we chose, the meta-explanation techniques we used, and the one we designed, accompanied by an ablation study. Additionally, we study the influence of the $noise$ and $\delta$ values on the evaluation, as well as we compare \textit{truthfulness} to another metric, \textit{complexity}.

\subsubsection{Truthfulness evaluation}
The first part of the experiments focuses on the evaluation of interpretations provided by linear neural models (NN1). Those are ground truth, and therefore we want to assess if the \textit{truthfulness} metric presented in Section~\ref{ssec:truth} correctly identifies these interpretations.

\begin{table}[ht]
\centering
\caption{Truthfulness evaluation of linear models (NN1) on the different datasets}
\label{tab:linear}
\begin{tabular}{ccccccccccccc}
\hline
         & \multicolumn{4}{c}{Weak}       & \multicolumn{4}{c}{Normal}     & \multicolumn{4}{c}{Strong}     \\ \hline
Dataset  & 0     & 0.0001 & 0.001 & 0.01  & 0     & 0.0001 & 0.001 & 0.01  & 0     & 0.0001 & 0.001 & 0.01  \\ \hline
TEDS     & 0.00  & 0.00   & 0.00  & 0.00  & 0.00  & 0.00   & 0.00  & 0.00  & 0.00  & 0.00   & 0.00  & 0.00  \\
CCA      & 0.00  & 0.00   & 0.00  & 0.00  & 0.00  & 0.00   & 0.00  & 0.00  & 0.00  & 0.00   & 0.00  & 0.00  \\
HDE      & 0.58  & 0.34   & 0.06  & 0.01  & 0.70  & 0.51   & 0.21  & 0.04  & 1.29  & 1.18   & 0.79  & 0.22 \\
MedN     & -     & -      & -     & -     & 0.00  & 0.00   & 0.00  & 0.00  & -     & -      & -     & -     \\ \hline
\end{tabular}
\end{table}

In Table~\ref{tab:linear}, we present the assessment of the inherent interpretation of NN1s on the different datasets. We use three different levels of $noise$ ($noise \in [$``weak'', ``normal'', ``strong''$]$), and four different $\delta$ values ($\delta \in [0, 0.0001, 0.001, 0.01]$). We know that NN1s are interpretable. Therefore, their interpretations must be 100\% truthful. In the results of the table, we can see that in TEDS (sensor level), CCA, and MedN, it perfectly evaluates the linear interpretations. However, in HDE, which is at the superpixel level, it assigns on average 1.29 wrong weights out of 29.71 superpixels. 

This is reasonable as in our implementation, if the value of a pixel is increased (decreased) above (below) the maximum (minimum) value after the alteration, then the value is reset to the maximum (minimum) value. Then, considering that the superpixels are working by averaging the per-pixel importance scores, if a pixel does not get increased (decreased) as the others, due to such a limitation, the possibility of observing unexpected behaviours increases. On the other hand, if we were conducting the experiment at the pixel level, that issue would not occur.

Let's see the following example. If we have 3 values $[0.8, 0.6, 0.9]$ (3 pixels in a superpixel) all of them having a range of $[0,1]$, with the importance weights $[0.2, 0.05, -0.35]$, and the prediction $= sigmoid(0.2\times0.8 + 0.05\times0.6 \times -0.35\times0.9) = sigmoid(-0.125) = 0.469$, the average importance would be $-0.04$ (the weight of the superpixel). If we make a positive alteration by $0.2$ the 3 values get $[1, 0.8, 1]$. Notice that $0.9$ changed to $1$ instead of $1.1$, in order to not violate the range $[0,1]$. The prediction will accordingly change to $= sigmoid(0.2\times1 + 0.05\times0.8 \times -0.35\times1) = sigmoid(-0.125) = 0.472$. While a positive alteration, given a negative weight, should lead to decreased prediction, this did not happen. If we had allowed the value $0.9$ to get $1.1$, the prediction will be $0.444$, hence decreased, as expected. We are aware of this limitation, but as it seems from the experiments is rare, and therefore, we decide to keep the restriction of keeping the values always inside their ranges.

Given these findings, we can conclude that when the interpretations are correct, \textit{truthfulness} accurately evaluates a technique. We shall put it to the test in non-linear, complex models in the following experiments.

\subsubsection{Interpretation techniques evaluation}
Let's examine how truthful the different interpretation techniques are on the four selected datasets. For the $noise$ and $\delta$ parameters, we choose ``normal'' and $0.001$ (default parameters), respectively. In Table~\ref{tab:NNeval}, for the four different datasets, we can observe how the different interpretation techniques correctly assign weights to the predictions of NN2 and NN3. Let's focus on the first four rows, which are the typical interpretation techniques, and their evaluations.

The two interpretation strategies that perform best in these cases are IG and LRP. There is no apparent winner between the two, since IG outperforms LRP in TEDS while performing similarly to CCA, and LRP outperforms IG in HDE and MedN. LIME, on the other hand, is the worst interpretation approach, doing worse than random interpretations in three of the four cases. 

\begin{table}[ht]
\centering
\caption{Avg. number of identified untruthful features per interpretation technique. Best is in bold, second best is underlined}
\label{tab:NNeval}
\begin{tabular}{ccccccccc}
\hline
       & \multicolumn{2}{c}{TEDS} & \multicolumn{2}{c}{CCA} & \multicolumn{2}{c}{HDE} & \multicolumn{2}{c}{MedN} \\ \hline
        & NN2         & NN3        & NN2        & NN3        & NN2        & NN3        & NN2         & NN3        \\ \hline
IG      & \underline{4.09}        & \underline{2.22}       & 4.45       & 4.59       & 18.09      & 18.09      & 4.31        & 5.16       \\
LRP     & 7.41        & 5.82       & 4.26       & 4.59       & \underline{15.35}      & \underline{17.36} & 4.17        & \underline{3.19}       \\
LIME    & 11.46       & 11.67      & 6.49       & 6.56       & 20.73      & 23.13      & 4.33        & 4.50       \\
Random  & 8.34        & 7.65       & 5.59       & 5.62       & 19.29      & 22.10      & 5.16        & 5.65       \\ \hline
Mean    & 7.77        & 6.96       & \underline{4.25}       & \underline{4.34}       & 17.18      & 19.21      & \underline{3.54}        & 3.74       \\
Median  & 7.41        & 6.29       & 4.45       & 4.59       & 17.12      & 18.45      & 3.74        & 3.83       \\
ourMeta & \textbf{2.87}        & \textbf{1.68}       & \textbf{1.36}       & \textbf{1.35}       & \textbf{11.70}      & \textbf{15.09}      & \textbf{0.50}        & \textbf{0.29}       \\ \hline
\end{tabular}
\end{table}

\subsubsection{Meta-explanation techniques comparison}
The performance of meta-explanation techniques, also known as ensembles, is shown in the same table (Table~\ref{tab:NNeval}). As we showed in the previous section, there is no clear winner among the many interpretation techniques in several cases. Given the need for a meta-explanation, the ensembling technique appears to be a good choice. Averaging the interpretation proves to be a promising approach based on the literature. Inspired by this, we included in the experiments a meta-explanation technique based on Median (selects the median weight among the suggested), and our meta-explanation technique as proposed in Section~\ref{sec:ourMeta}. 

Checking the results from Table~\ref{tab:NNeval}, we can see that our meta-explanation technique can drastically reduce the number of identified untruthful features, in all cases. All three meta-explanation techniques use the four interpretation techniques as input. Our technique reduces the untruthful features by 63\% compared to the original techniques, and by 54\% compared to the other two meta-explanation techniques.

\paragraph{Ablation study.} We will also conduct an ablation study on seed interpretation techniques. We will remove each technique one at a time, monitoring how the meta-explanation techniques perform. In Table~\ref{tab:ablation}, we can see the results. The performance of the meta-explanation techniques employing all four interpretation techniques is shown in the first row. In the next rows, we omit the following approaches in this order: IG, LRP, LIME, and Random. The red up arrow (\ruparr) indicates that omitting the specific interpretation technique reduced the meta-explanation technique's performance compared to the original performance. The green down arrow (\gdoarr) indicates that performance improved as the average number of untruthful elements in meta-explanations decreased.

\begin{table}[ht]
\centering
\caption{Ablation study regarding the meta-explanation techniques}
\label{tab:ablation}
\begin{tabular}{cccccccccc}
\hline
     &  & \multicolumn{2}{c}{TEDS} & \multicolumn{2}{c}{CCA} & \multicolumn{2}{c}{HDE} & \multicolumn{2}{c}{MedN} \\
           &        &   NN2          &   NN3        &    NN2         &    NN3        &    NN2         &     NN3        & NN2          & NN3         \\ \hline
           & Mean   & 7.77           & 6.96         & 4.25           & 4.34          & 17.18          & 19.21          & 3.54         & 3.74        \\
Original   & Median & 7.41           & 6.29         & 4.45           & 4.59          & 17.12          & 18.45          & 3.74         & 3.83        \\
           & Ours   & 2.87           & 1.68         & 1.36           & 1.35          & 11.70          & 15.09          & 0.50         & 0.29        \\ \hline
           & Mean   & 9.25\ruparr    & 8.70\ruparr  & 4.90\ruparr    & 5.00\ruparr   & 17.36\ruparr   & 20.17\ruparr   & 3.73\ruparr  & 3.71\ruparr \\
w/o IG     & Median & 9.41\ruparr    & 8.67\ruparr  & 5.25\ruparr    & 5.43\ruparr   & 17.96\ruparr   & 20.37\ruparr   & 4.24\ruparr  & 3.91\ruparr \\
           & Ours   & 4.67\ruparr    & 3.31\ruparr  & 1.53\ruparr    & 1.52\ruparr   & 12.13\ruparr   & 15.37\ruparr   & 0.90\ruparr  & 0.59\ruparr \\ \hline
           & Mean   & 7.81\ruparr   & 7.81\ruparr   & 4.91\ruparr    & 4.90\ruparr   & 18.90\ruparr   & 20.52\ruparr   & 3.89\ruparr  & 4.26\ruparr \\
w/o LRP    & Median & 7.74\ruparr   & 6.85\ruparr   & 5.38\ruparr    & 5.41\ruparr   & 19.39\ruparr   & 20.71\ruparr   & 4.34\ruparr  & 4.81\ruparr \\
           & Ours   & 3.05\ruparr   & 1.77\ruparr   & 1.51\ruparr    & 1.56\ruparr   & 12.90\ruparr   & 15.63\ruparr   & 1.07\ruparr  & 1.21\ruparr \\ \hline
           & Mean   & 6.65\gdoarr   & 5.51\gdoarr   & 4.40\ruparr    & 4.30\gdoarr   & 16.72\gdoarr   & 18.75\gdoarr   & 3.99\ruparr  & 4.36\ruparr \\
w/o LIME   & Median & 5.78\gdoarr   & 6.85\ruparr   & 4.38\gdoarr    & 4.55\gdoarr   & 16.68\gdoarr   & 17.78\gdoarr   & 4.16\ruparr  & 4.19\ruparr \\
           & Ours   & 2.95\ruparr   & 1.73\ruparr   & 2.60\ruparr    & 2.48\ruparr   & 12.32\ruparr   & 15.47\ruparr   & 1.02\ruparr  & 0.64\ruparr \\ \hline
           & Mean   & 7.52\gdoarr   & 6.53\gdoarr   & 4.05\gdoarr    & 3.98\gdoarr   & 16.85\gdoarr   & 18.51\gdoarr   & 3.45\gdoarr  & 3.57\gdoarr \\
w/o Random & Median & 7.08\gdoarr   & 5.36\gdoarr   & 4.48\ruparr    & 4.66\ruparr   & 17.22\ruparr   & 17.92\gdoarr   & 3.92\ruparr  & 3.64\gdoarr \\
           & Ours   & 3.00\ruparr   & 1.75\ruparr   & 2.37\ruparr    & 2.39\ruparr   & 12.66\ruparr   & 15.60\ruparr   & 0.92\ruparr  & 0.58\ruparr \\ \hline

\end{tabular}
\end{table}

When LIME or Random are omitted, our meta-explanation technique appears to perform slightly worse, whereas the other two perform slightly better than their original performance. This occurs as a result of erroneous interpretations introducing $noise$ into these two meta-explanation techniques. Our meta-explanation technique, on the other hand, detects the correct elements even in these interpretations and uses only them, discarding any potentially noisy ones.

Another intriguing discovery is that the Mean and Median meta-explanation techniques are susceptible to changes in the seed interpretation techniques. On average, the performance of both techniques changes $0.60$ and $0.77$. Contrarily, our meta-explanation technique appears to be more stable. On both neural networks, the average change across all datasets is $0.56$. We assume that adding more seed interpretation techniques will significantly enhance these observations.

\subsubsection{Parameters impact}
In the prior experiments, we identified IG as the best interpretation technique, and our meta-explanation technique as the best across the other meta-explanation techniques. We select these two and LIME, which is probably the most popular interpretation technique, to analyse how they perform given different $noise$ and $\delta$ values in our cases. 

\begin{table}[ht]
\centering
\caption{Performance of IG, LIME and Ours meta-explanation technique on \textit{truthfulness} based on different $noise$ and $\delta$ values}
\label{tab:noise_and_delta_values}
\begin{tabular}{cccccccccccccc}
\hline
      &         & \multicolumn{4}{c}{Weak}    & \multicolumn{4}{c}{Normal}    & \multicolumn{4}{c}{Strong}    \\
      &         & 0     & 0.0001 & 0.001 & 0.01    & 0      & 0.0001  & 0.001 & 0.01   & 0     & 0.0001 & 0.001 & 0.01 \\ \hline
      & IG      & 3.5   & 3.4    & 2.9   & 1.1     & 4.2    & 4.1     & 3.8   & 2.3    & 4.8   & 4.8    & 4.5   & 3.6  \\
TEDS  & LIME    & 11.0  & 10.9   & 10.0  & 6.0     & 11.5   & 11.5    & 11.0  & 8.6    & 12.0  & 12.0   & 11.8  & 10.4 \\
      & Ours    & 2.2   & 2.1    & 1.8   & 0.6     & 2.9    & 2.8     & 2.6   & 1.5    & 3.8   & 3.7    & 3.6   & 2.6  \\ \hline
            
      & IG      & 4.3   & 4.2    & 3.9   & 2.6     & 4.5    & 4.5     & 4.3   & 3.2    & 4.8   & 4.8    & 4.7   & 3.9  \\
CCA   & LIME    & 6.2   & 6.3    & 5.9   & 3.8     & 6.5    & 6.5     & 6.3   & 4.8    & 6.6   & 6.6    & 6.5   & 5.5  \\
      & Ours    & 1.1   & 1.0    & 0.9   & 0.5     & 1.4    & 1.4     & 1.2   & 0.8    & 1.8   & 1.7    & 1.7   & 1.2  \\ \hline
                
      & IG      & 17.6  & 15.3   & 10.9  & 4.0     & 19.7   & 18.1    & 14.5  & 7.4    & 20.6  & 19.6   & 16.9  & 10.2 \\
HDE   & LIME    & 20.7  & 18.4   & 13.2  & 4.8     & 22.3   & 20.7    & 16.8  & 8.4    & 23.2  & 22.2   & 19.3  & 11.7 \\
      & Ours    & 9.4   & 7.8    &  4.9  & 1.3     & 13.0   & 11.7    & 8.9   & 3.7    & 15.3  & 14.3   & 11.8  & 6.2  \\ \hline
        
      & IG      &   -   &   -    &   -   &    -    & 21.2   & 4.3     &  1.9  & 0.6    &   -   &   -    &   -   &   -  \\
MedN  & LIME    &   -   &   -    &   -   &    -    & 22.9   & 4.3     &  1.8  & 0.5    &   -   &   -    &   -   &   -  \\
      & Ours    &   -   &   -    &   -   &    -    &  2.6   & 0.5     &  0.2  & 0.1    &   -   &   -    &   -   &   -  \\ \hline
\end{tabular}
\end{table}

In Table~\ref{tab:noise_and_delta_values}, we can see the performance of the techniques given the different $noise$ and $\delta$ values changes on the NN2 neural network. In every case, ``weak'' $noise$ and higher $\delta$ values (e.g., $0.01$) produce higher \textit{truthfulness} scores and do not allow to easily distinguish between techniques. On the other hand, ``strong'' $noise$ and $\delta=0$ is very strict and punitive. Therefore, we suggest the use of ``normal'' $noise$ with a small $\delta=0.0001$, but not $0$. 

\begin{table}[ht]
\centering
\caption{Comparison of \textit{truthfulness} and \textit{complexity} between IG, LIME and Ours meta-explanation technique with different $\delta$ values}
\label{tab:nzw_comparison}
\begin{tabular}{cccccccccccccccccc}
\hline
      &      & \multicolumn{2}{c}{0} & \multicolumn{2}{c}{0.0001} & \multicolumn{2}{c}{0.001} & \multicolumn{2}{c}{0.01}\\
      &      & Tr     & Co    & Tr     & Co    & Tr    & Co    & Tr   & Co    \\ \hline
      & IG   & 4.2    & 14.0  & 4.1    & 13.98 & 3.8   & 13.72 & 2.3  & 11.43 \\
TEDS  & LIME & 11.5   & 14.0  & 11.5   & 13.98 & 11.0  & 13.78 & 8.6  & 11.89 \\
      & Ours & 2.9    & 11.08 & 2.8    & 11.11 & 2.6   & 11.12 & 1.5  & 9.19  \\ \hline
      & IG   & 4.5    & 10.28 & 4.5    & 10.28 & 4.3   & 10.26 & 3.2  & 10.03 \\
CCA   & LIME & 6.5    & 11.0  & 6.5    & 10.99 & 6.3   & 10.97 & 4.8  & 10.71 \\
      & Ours & 1.4    & 10.62 & 1.4    & 10.62 & 1.2   & 10.71 & 0.8  & 10.64 \\ \hline
      & IG   & 19.7   & 29.71 & 18.1   & 29.54 & 14.5  & 28.03 & 7.4  & 15.40 \\
HDE   & LIME & 22.3   & 26.05 & 20.7   & 25.98 & 16.8  & 25.37 & 8.4  & 19.08 \\
      & Ours & 13.0   & 16.67 & 11.7   & 17.74 & 8.9   & 18.45 & 3.7  & 7.63  \\ \hline
      & IG   & 21.2   & 48.25 & 4.3    & 48.13 & 1.9   & 47.17 & 0.6  & 38.27 \\
MedN  & LIME & 22.9   & 44.75 & 4.3    & 44.74 & 1.8   & 44.62 & 0.5  & 43.32 \\
      & Ours & 2.6    & 45.41 & 0.5    & 41.63 & 0.2   & 32.88 & 0.1  & 18.01 \\ \hline
\end{tabular}
\end{table}

\subsubsection{Comparison with another metric}

One last quantitative experiment compares the \textit{truthfulness} metric to another well-known interpretability metric known as \textit{complexity}. \textit{Complexity} measures the number of non-zero weights included in an interpretation. Lower scores in this metric suggest lower complexity, which means more comprehensible explanations. We can include a \textit{complexity} threshold. The importance scores that fall below that threshold are then regarded zero, and the number of non-zero elements is reduced. In the results shown in Table~\ref{tab:nzw_comparison}, where we use ``normal'' $noise$, we use the $\delta$ values as the \textit{complexity} threshold as well. For $\delta=0.0001$, the evaluation of the different techniques is unclear, and we cannot easily choose the best technique. For example, in TEDS dataset, both IG and LIME have the same \textit{complexity}, but they have very different \textit{truthfulness} scores, with LIME making twice the mistakes as IG. 

Another intriguing finding from the \textit{complexity} metric is that our meta-explanation technique, which always shows the highest \textit{truthfulness} score, also has the lowest \textit{complexity} scores in three of the four test cases. This means that the meta-explanation technique can reduce the number of elements that appear to the end user, making explanations shorter and easier to understand, while guaranteeing that the remaining importance scores are truthful. This is due to the meta-explanation technique just replacing the incorrect components with zero. Given that the \textit{truthfulness} score is always lower, this replacement is most likely correct. As a result, it not only ensembles but also corrects the seed interpretation techniques.





\subsection{Qualitative experiments}
\label{sec:argExa}

In this section, we will present two examples comparing interpretations provided by IG, LRP, and Mean, with our meta-explanation technique from the textual (MedN) and image (HDE) datasets. In both cases, we use the NN3 neural model. Moreover, we will showcase how the argumentation framework can be employed to provide richer explanations.

\begin{figure}[ht]
  \centering
  \includegraphics[width=1\linewidth]{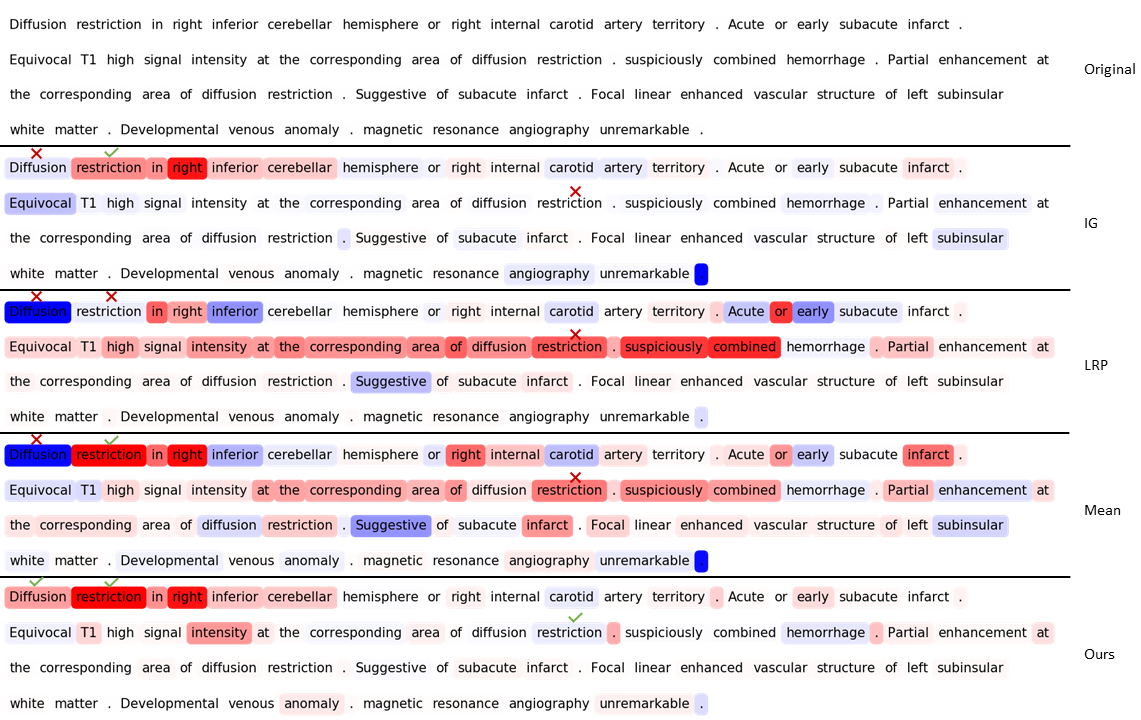}
  \caption{MedN Example: Explanation provided by different interpretation techniques for a specific instance. The colour red indicates that the word has a positive influence on the prediction, whereas the colour blue suggests that it has a negative effect. Green mark indicates that the influence is correct, while red cross incorrect}
  \label{fig:first_example}
\end{figure}

We will start with the first example regarding an instance from the MedN dataset. In the first row in Figure~\ref{fig:first_example}, we can see the examined instance. The prediction from the neural network regarding this medical report was 93\% probability to concern acute ischemic stroke. Focusing on few specific words, ``Diffusion'', ``Restriction'' (appearing 1st), ``Restriction'' (appearing 2nd). All these words have been assigned with a truthful weight from our meta-explanation technique, and the corresponding arguments are presented below. For example, regarding the word ``Diffusion'', which according to IG, LRP and Mean, should have a negative weight, when it is removed, the probability drops. Therefore, the weight should have been positive, as our meta-explanation technique correctly assigned. The corresponding argument is the $f_{Diffusion,DEC}$.

\begin{itemize}
    \item []$f_{Diffusion,DEC}$ The evaluation of the alteration of $Diffusion$'s value $1$ to $0$ ($DEC$) was performed and the model's behaviour was as expected $DEC$ (93\% to 90\%), according to its importance $z_{Diffusion}=0.75$.

    \item []$f_{Restriction-1st,DEC}$: The evaluation of the alteration of $Restriction-1st$'s value $1$ to $0$ ($DEC$) was performed and the model's behaviour was as expected $DEC$ (93\% to 86\%), according to its importance $z_{Restriction-1st}=0.92$.

    \item []$f_{Restriction-2nd,INC}$: The evaluation of the alteration of $Restriction-2nd$'s value $1$ to $0$ ($INC$) was performed and the model's behaviour was as expected $INC$ (93\% to 95\%), according to its importance $z_{Restriction-2nd}=-0.29$.
\end{itemize}

\begin{figure}[ht]
  \centering
  \includegraphics[width=0.95\linewidth]{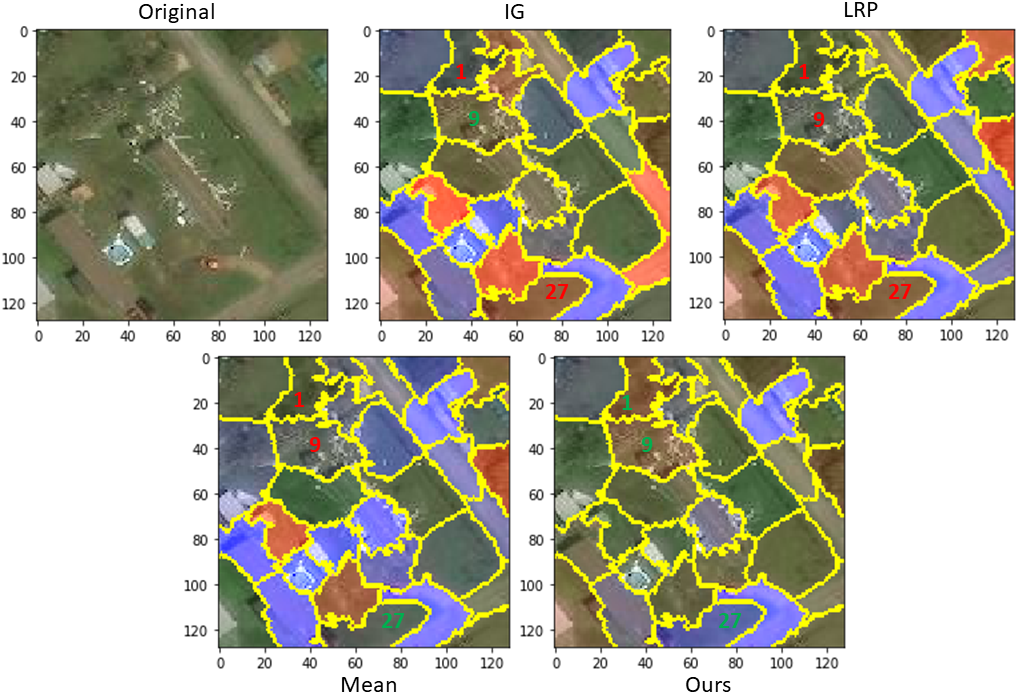}
  \caption{CCA Example: Explanation provided by different interpretation techniques for a specific instance. The colour red indicates that the word has a positive influence on the prediction, whereas the colour blue suggests that it has a negative effect. Green numbers indicates that the weight of the segment is correct, while red numbers incorrect}
  \label{fig:second_example}
\end{figure}

In the second example (Figure~\ref{fig:second_example}), we can see an examined instance, which is classified as ``damaged'' with a 60\% probability. We can see the different interpretations from the techniques, at the segment (superpixel) level. Red highlight indicates that the segment is very crucial to the prediction of this class, while blue for the other class. Our meta-explanation method achieves to both highlight fewer parts of the image, while it contains less untruthful weights. The segments, with number 1, 9 and 27, are presented. The weights assigned to all of them is correct in our meta-explanation, in contrast to the others. Segment 1 is negative to IG, LRP and Mean, while it should have been positive. The arguments supporting this are $f_{Segment_1,DEC}$ and $f_{Segment_1,INc}$. Similarly, to 9 and 27.

\begin{itemize}
    \item []$f_{Segment_1,DEC}$ The evaluation of the alteration of $Segment_1$'s value by $-0.07$ ($DEC$) was performed and the model's behaviour was as expected $DEC$ (60.5\% to 60.1\%), according to its importance $z_{Segment_1}=0.20$.
    \item []$f_{Segment_1,INC}$ The evaluation of the alteration of $Segment_1$'s value by $+0.0.07$ ($INC$) was performed and the model's behaviour was as expected $INC$ (60.5\% to 60.9\%), according to its importance $z_{Segment_1}=0.20$.

    \item []$f_{Segment_9,DEC}$ The evaluation of the alteration of $Segment_9$'s value by $-0.26$ ($DEC$) was performed and the model's behaviour was as expected $DEC$ (60.5\% to 39.2\%), according to its importance $z_{Segment_9}=0.24$.
    \item []$f_{Segment_9,INC}$ The evaluation of the alteration of $Segment_9$'s value by $+0.26$ ($INC$) was performed and the model's behaviour was as expected $INC$ (60.5\% to 62.7\%), according to its importance $z_{Segment_9}=0.24$.
        
    \item []$f_{Segment_27,DEC}$ The evaluation of the alteration of $Segment_27$'s value by $-0.14$ ($DEC$) was performed and the model's behaviour was as expected $INC$ (60.5\% to 62.5\%), according to its importance $z_{Segment_27}=-0.75$.
    \item []$f_{Segment_27,INC}$ The evaluation of the alteration of $Segment_27$'s value by $+0.14$ ($INC$) was performed and the model's behaviour was as expected $DEC$ (60.5\% to 52.3\%), according to its importance $z_{Segment_27}=-0.75$.
    
\end{itemize}

We believe that by presenting such arguments to the user, we may help them evaluate and choose the optimal method from among multiple options, as well as enhance their faith in the system. Specifying the reasons why an importance score is regarded truthful or untruthful can considerably increase trust in the interpretation.

\section{Conclusions}
\label{sec:concl}

Machine learning models must be interpretable when used in high-risk applications. There are several interpretability techniques for explaining a model's decisions, as well as evaluation methods for determining the quality of the explanations. However, because there are so many alternatives, determining the best interpretation technique for a given application can be challenging. While evaluation can assist in this process, it is not always the case. It would be extremely beneficial to give a user-friendly evaluation metric that would allow the combining of various interpretation techniques via a meta-explanation framework.

Truthfulness is an excellent choice for a meta-explanation technique, whereas alternative faithfulness-based metrics produce outputs that would be difficult to incorporate into a local ensembling/meta-explanation technique. Truthfulness is appropriate since it filters the important scores of each interpretation if they have the incorrect polarity based on the model's behaviour with a few alterations. Depending on this filtering, the meta-explanation technique may readily select the most appropriate truthful score among the scores based on the change in the output. As a result, with fully unsupervised methods, both interpretation techniques and metrics, the meta-explanation we provide is an excellent choice.

We investigated the ability of \textit{truthfulness} to appropriately assess interpretations, intrinsic or not, in four datasets of diverse data types through large-scale experimentation. Then, we discussed the performance of meta-explanation techniques while conducting an ablation study, demonstrating that our meta-explanation technique always performs better when the number of input (seed) interpretations is increased, even when noisy interpretations are included, whereas other methods perform better when noisy or erroneous interpretations are omitted. 

Although we used IG, LIME, and LRP in our experiments, our technique is not limited to them. To improve the performance of the meta-explanation technique, the end user can easily replace or add new ones. Nonetheless, they must always keep in mind that the additional interpretation techniques must be consistent with the two assumptions (Assumption~\ref{assumption:1} and \ref{assumption:2}).

A discussion regarding the two parameters of \textit{truthfulness}, $noise$ and $\delta$, also took place, to let users decide based on their applications which values are the most appropriate. For a stricter evaluation, a ``strong'' $noise$ with a $0$ $\delta$ value would be suggested. However, in most cases, ``normal'' $noise$ with $0.0001$ $\delta$ value would be enough. We suggest this value as such difference in the prediction is almost insignificant in most applications.

Through large-scale experimentation, we explored the ability of \textit{truthfulness} to accurately assess interpretations, intrinsic or not, in four datasets of varied data types. Then, while conducting an ablation study, we discussed the performance of meta-explanation techniques, demonstrating that our meta-explanation technique always performs better when the number of input (seed) interpretations is increased, even when noisy interpretations are included, whereas other methods perform better when noisy or erroneous interpretations are omitted. This allows us to freely add interpretations as seeds in our meta-explanation, considering solely the additional computational overhead as it improves efficiency while overlooking potential noise.

A debate about the two \textit{truthfulness} parameters, $noise$ and $\delta$, took place as well, to allow users to select which values are most appropriate for their applications. A ``strong'' $noise$ with a $\delta$ value of $0$ is recommended for a more stringent evaluation. However, in most circumstances, ``normal'' $noise$ with a $\delta$ value of $0.0001$ would suffice. In most situations, such difference in prediction is almost minor, hence we recommend this setting. We also compared \textit{truthfulness} to the \textit{complexity} metric. Based on this comparison, we wanted to highlight that our meta-explanation technique almost always delivers both the most truthful and the least complex interpretations, especially as the $\delta$ value increases.

Finally, we demonstrated a qualitative experiment using two examples from two distinct datasets. We contrast our meta-explanation technique with various interpretability techniques. Our meta-explanation technique is less complex and more truthful. The \textit{truthfulness} of each important score is additionally supported by a few arguments, which assist the end user trust the interpretation. Those arguments are re-phrased compared to the originals, as presented in our preliminary work~\cite{altruist}.

Future objectives include conducting a large-scale experiment to compare \textit{truthfulness} to other metrics and determining whether or not there is a correlation between them. In order to provide even better evaluations, we will also look into ways to increase \textit{truthfulness}. To capture the polarity more accurately within a given range of values, we would also like to investigate replacing the two alterations we perform for each feature value with an interpolation approach. Additional machine learning models, like for example Transformers, can be examined, as well as different tasks, like multi-class and multi-label classification. In addition, we intend to improve the argumentation framework and the meta-explanation technique. Finally, we would like to conduct a human-centred experiment to demonstrate that end users prefer meta-explanation techniques.

\section*{Acknowledgment}
This work has been supported by the European Union's Horizon 2020 research and innovation programme under grant agreement No. 825619 (AI4EU~\footnote{\url{https://www.ai4europe.eu}}).

\bibliographystyle{unsrt}

\end{document}